\documentclass[conference]{IEEEtran}
\IEEEoverridecommandlockouts
\usepackage{cite}
\usepackage{amsmath,amssymb,amsfonts}
\usepackage{algorithmic}
\usepackage{graphicx}
\usepackage{textcomp}
\usepackage{xcolor}
\usepackage{float} 
\usepackage{hyperref}
\def\BibTeX{{\rm B\kern-.05em{\sc i\kern-.025em b}\kern-.08em
    T\kern-.1667em\lower.7ex\hbox{E}\kern-.125emX}}

\definecolor{purple}{cmyk}{0,0.98,0.41,0.3}
\definecolor{myyellow}{cmyk}{0,.15,1,0}

\begin{document}

\title{Semantic analysis of behavior in a DNA-functionalized molecular swarm}

\author{\IEEEauthorblockN{Tom Bachard}
\IEEEauthorblockA{\textit{Ochanomizu University}\\
Tokyo, Japan}
\and
\IEEEauthorblockN{Gong Yiming}
\IEEEauthorblockA{\textit{University of Kyoto}\\
Kyoto, Japan}
\and
\IEEEauthorblockN{Ibuki Kawamata}
\IEEEauthorblockA{\textit{University of Kyoto}\\
Kyoto, Japan}
\and
\IEEEauthorblockN{Akira Kakugo}
\IEEEauthorblockA{\textit{University of Kyoto}\\
Kyoto, Japan}
\and
\IEEEauthorblockN{Nathanaël Aubert-Kato}
\IEEEauthorblockA{\textit{Ochanomizu University}\\
Tokyo, Japan}
}

\maketitle

\begin{abstract}
In this paper, we propose applying semantic embedding to learn the range of behaviors exhibited by molecular swarms, thereby providing a richer set of features to optimize such systems. Specifically, we consider a standard molecular swarm where the individuals are cytoskeletal filaments (called microtubules) propelled by surface-adhered kinesin motors, with the addition of DNA functionalization for further control. We extend a microtubule model with that additional interaction and show that the extracted semantic atoms from simulation results match the expected behaviors. Moreover, the decomposition of each frame in the simulations accurately describes the expected impact of the external control values. Those results provide relevant leads towards the explainability of simulated experiments, making them more reliable for designing and optimizing in-vitro systems.
\end{abstract}

\section{Introduction}

In the natural world, large collections of agents, be they ants, birds, or fish, can self-organize into massive structures capable of solving problems far beyond the capabilities of a single individual. That phenomenon has been the inspiration for the field of swarm robotics \cite{dorigo2020reflections,floreano2021individual}, in which limited agents (robots) interact locally with each other in order to form structures greatly exceeding the scale of the individual.
Theoretical work has shown that the complexity of such structures will increase with the size of the swarm, providing an incentive to aim for millions of units, or more \cite{witkowski2019make}. One approach to gather a swarm of that size is to go to the micrometer scale, where robots are made of molecular structures interacting with each other through chemical reactions \cite{hagiya2014molecularrobots,kabir2020molecular}, demonstrating swarms of millions \cite{aubert2017swarmrobots,zadorin2017synthesis} or even hundreds of millions of units \cite{KeyaKakugo2018,kawamata2024autonomous}.

However, that scaling factor comes at a cost: molecular robots are much harder to program than their electronic counterparts. A typical molecular controller will rely on the concentration of specific molecules to encode data and chemical reactions among those molecules to implement computational processes \cite{hagiya2014molecularrobots}. The non-linearity and inherent parallelism of the operations make the rational design of such systems unrealistic beyond simple toy applications. To push past that stage, one may either use high-throughput experimental setups to sample the search space \cite{genot2016high,baccouche2017massively}, rely on optimization algorithms to find (often unexpected) solutions \cite{aubert2017swarmrobots,cazenille2019exploring}, or both \cite{gutierrez2014evolution,henson2015towards}. Regardless of the method, that design step ends up generating thousands or even millions of experimental results, far more than can be manually checked by a human experimentalist. As such, the automated extraction of information becomes a necessary step to effectively use that trove of data.

While global features (\emph{e.g.}, concentration of a given molecule indicating the completion of a task, local alignment between robots, convex hull of the swarm, and so on) may be enough for optimization, they do not capture high-level information about the behavior of the swarm. In this paper, we propose to apply semantic embedding to learn the range of behaviors of a molecular swarm, providing a richer set of features to optimize such systems. Here, the semantic embedding refers to the mapping of the abstract concept of the semantics of the data to low-dimensional numerical vectors expressing those semantics as combinations of generic concepts (the basis of a semantic space). These semantic spaces are usually obtained through the use of foundation models learning from a wide range of typical data.

Foundations models are multi-modal deep learning models that embed their input data into a latent space organized around a high-level description of the data (here, \textit{semantics}). %
Recently, CLIP (Contrastive Language-Image Pre-training)~\cite{clip}, a popular foundation model, showed promising high-level description and encapsulation of the semantic of natural images~\cite{bachard2022clip}. %
More precisely, this model is the cornerstone of a pipeline that is capable of summarizing the most redundant semantic components of an image collection~\cite{bachard2024smic} into a semantic dictionary. %
This property motivated us to apply this semantic approach to extract and study the semantic behaviors of molecular swarms. 

As a proof of concept, we focus here on videos of a simulated swarm of DNA-functionalized microtubules \cite{KeyaKakugo2018}. %
Visualization tools and statistical evaluation of the decomposition of the frames into a semantic dictionary confirm that the approach highlights and quantifies the behavior expected by experimentalists. %
As such, this work can be considered as a first step in the field of explainability for molecular swarms, providing higher-level features for further applications such as automated optimization.

In Section~\ref{sec:model} we introduce the experimental context of this work: the simulator -- based on C-GLASS~\cite{cglass} --, the model for DNA functionalization, and a basic categorization \textit{in silico} microtubule swarms. %
Then, in Section~\ref{sec:embedding}, we introduce, interpret, and visualize the semantic embedding (here a semantic dictionary \cite{bachard2024smic}), applied to such systems. %
Lastly, in Section~\ref{sec:interpretation} we quantify the extent to which  the observed semantic behaviors represent the simulated experiments, and we link these behaviors to the response of the swarm to a global parameter (here, temperature).

\section{Molecular swarm: DNA-functionalized Microtubules}
\label{sec:model}

\begin{figure}[htbp]
    \centering
    \includegraphics[width=\linewidth]{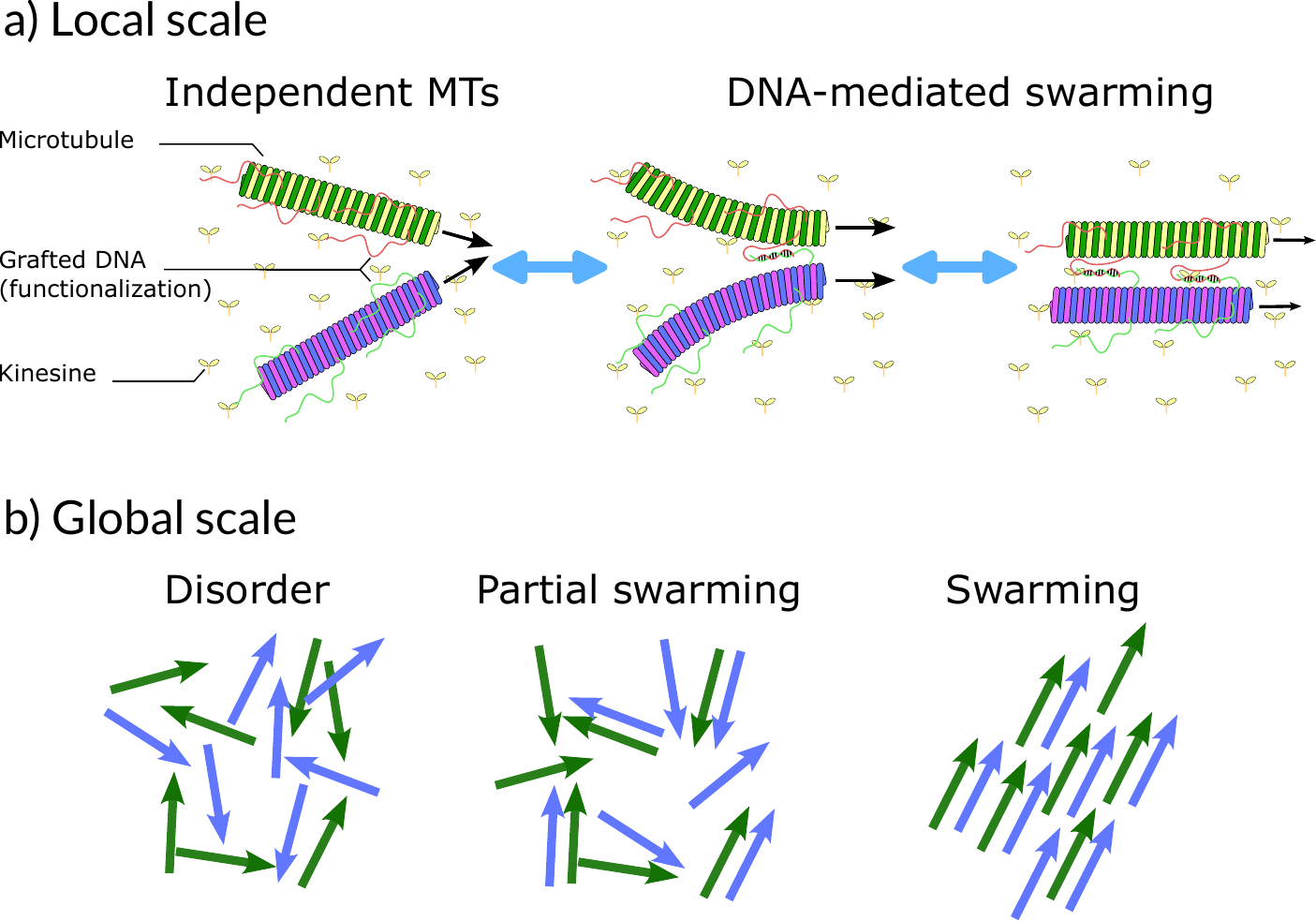}
    \caption{Molecular swarm. (a) At the local scale, microtubules (MTs) are carried over a bed of kinesin grafted to a glass surface. Upon collision, MTs will change their orientation. While such interaction is weak in regular MTs, we use DNA functionalization to enhance the swarming effect: complementary DNA strands attached to the surface of the MTs form duplexes, providing additional support to align the strands. Finally, those duplexes can be destabilized by an external output (here, temperature), reversing the process. (b) At the global scale, depending on the stability of DNA duplexes, we observe three types of behaviors: disorder, when duplexes are unstable; partial swarming, where groups can form with MTs continuously joining and leaving; swarming, when large groups recruit many MTs, with nearly no disassembly.}
    \label{fig:dna_model}
\end{figure}

In this paper, we consider a molecular swarm implemented by a microtubule–kinesin active matter system \cite{Surrey2001,Sanchez2012,Dogic2019}. Microtubules are tubulin polymers that provide structural functions and, in the cell, form ``paths'' for molecular motors like kinesin. The active matter approach, however, flips this configuration around: kinesin molecules are densely fixed to a glass slide, forming a disordered conveyor belt for short microtubules (Figure~\ref{fig:dna_model}, top). In that context, microtubules are seen as the active units of the swarm rather than being a passive path like in living systems. Furthermore, we consider that microtubules have been functionalized with artificial DNA strands. DNA strands provide a way to selectively connect microtubules to each other, as complementary DNA sequences will form duplexes (a process called hybridization), while incompatible sequences will not interact. Once connected, microtubules will be forced to align, a phenomenon that can be externally controlled through inputs reversing the duplexing \cite{KeyaKakugo2018,kawamata2024autonomous}. While collisions between microtubules already provide a simple alignment mechanism, the addition of DNA functionalization greatly increases the level of control available.

\subsection{Interactions mechanisms}

One of the major issues with training large representation models is the need for massive amounts of data. Here, we alleviate that issue by generating simulation results through a state-of-the-art simulator, C-GLASS \cite{cglass}. In particular, C-GLASS provides a model for the motility of microtubules on a bed of kinesin, considering the forces of propulsion, structural constraints, as well as interactions between microtubules. Specifically, the model considers microtubules to be self-propelled inextensible connected segments that are subjected to a bending potential (putting stress to keep the microtubule straight) and Lennard-Jones interaction forces \cite{moore2020collective}. In the model, forces are applied at the end of the segments, called sites. Formally, a microtubule is considered to be a list $(a_i)_{i\in\{1..N\}}$ of $N$ sites, corresponding to $N-1$ segments. The position $\mathbf{r}_i(t)$ of the $i$th site is computed using the midstep algorithm:

\begin{align}
    &\mathbf{r}_i(t+1/2) = \mathbf{r}_i(t) + \dfrac{\Delta t}{2} \mathbf{v}_i(t)\\
    &\mathbf{r}_i(t+1) = \mathbf{r}_i(t)+\Delta t \, \mathbf{v}_i(t+1/2)
\end{align}

where $\Delta t$ is the computational time step, $\mathbf{v}_i(t)$ is the velocity vector of site $i$ at time $t$, and $\mathbf{v}_i(t+1/2)$ the velocity vector at midstep recomputed from $\mathbf{r}_i(t+1/2)$.

The computation of the velocity vector is approximated from the forces applied at the site positions and a friction tensor:

\begin{equation}
    \mathbf{v}_i(t) = \zeta^{-1}_i(t) \cdot F^{tot}_i(t)
\end{equation}

\begin{figure*}[htbp]
    \centering
    \includegraphics[width=\linewidth]{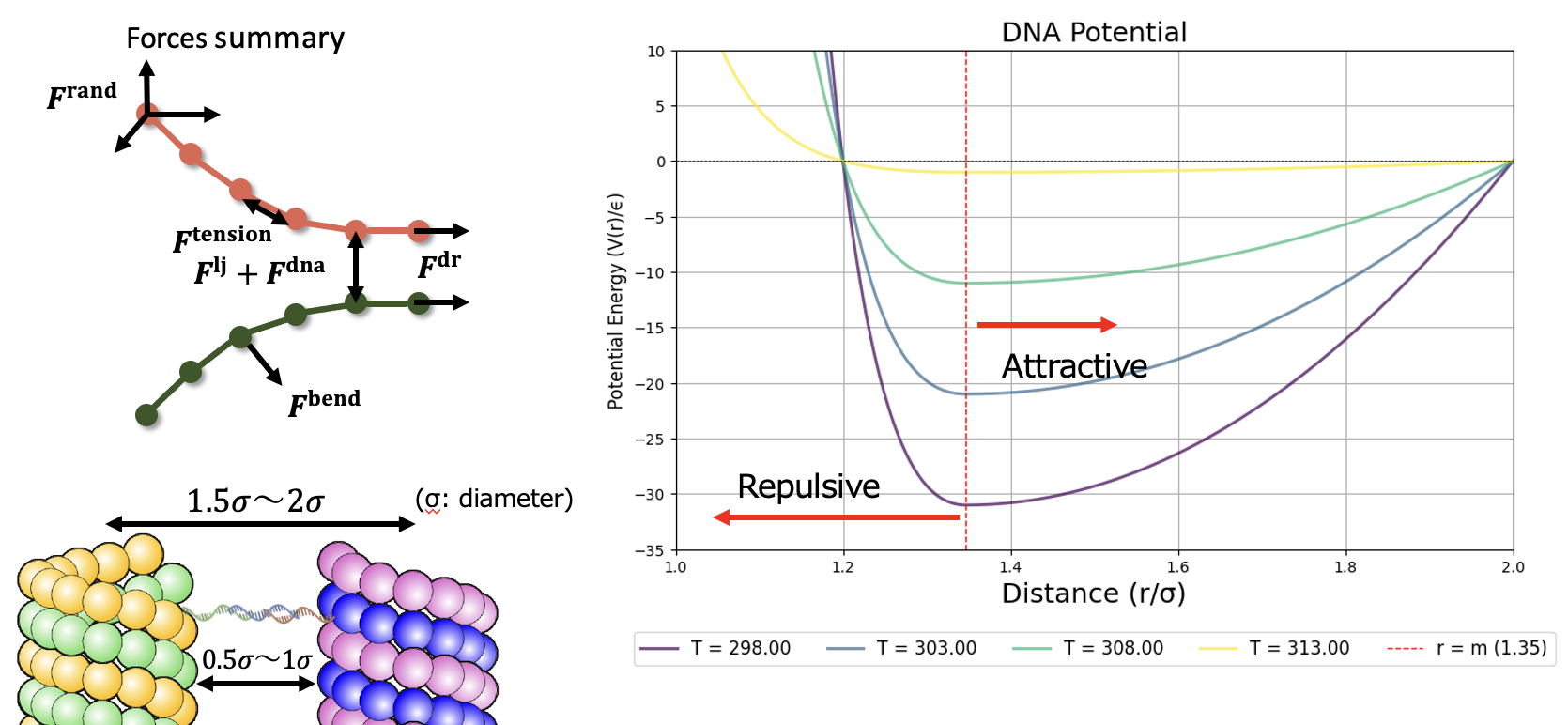}
    \caption{Forces applied to the microtubules. Forces based on a Lennard-Jones potential $F^{lj}$ corresponds to microtubule-microtubule interactions, while forces based on the DNA potential $F^{dna}$ corresponds to the strength of grafted DNA duplexing interactions. When the temperature increases, that interaction becomes weaker, according to the stability of the structure.}
    \label{fig:forces}
\end{figure*}

That model is further extended to introduce interactions between DNA molecules that were attached to the surface of the microtubules (Figure \ref{fig:forces}). As such, the computation of the forces applied to site $i$ at time $t$ becomes:

\begin{equation}
\begin{split}
    F^{tot}_i(t) = F^{bend}_i(t) + F^{tension}_i(t) + F^{dr}_i(t) \\
    + F^{lj}_i(t) + F^{dna}_i(t) + F^{rand}_i(t)
\end{split}
\end{equation}

where $F^{bend}_i$ is the force applied from the bend at site $i$,  $F^{tension}_i$ the force applied by neighboring segments, $F^{dr}_i$ the driving force applied by the kinesin, $F^{lj}_i(t)$ the Lennard-Jones interaction from nearby microtubules, $F^{rand}_i$ a random noise with 0 mean applied to the site, all computed in accordance with the previously established model \cite{moore2020collective}. $F^{dna}_i$ is our contribution, corresponding to a second-order potential established to represent the time DNA strands duplex with each other, thus impacting the forced alignment between microtubules by adding a spring force between sites:

\begin{align}
    U^{dna}(r) &= \dfrac{\epsilon}{(2-m)^2}(r-m)^2-\epsilon\label{eq:potential}\\
    m &= 1.5 \sigma\\
    \epsilon &= \Delta G(T) = \Delta H - T \Delta S
\end{align}

Where $\sigma$ is the diameter of the microtubules, $m$ is a parameter corresponding to the optimal duplexing length (based on the length of the DNA sequence), set here to $1.5\sigma$ based on size considerations (see Figure~\ref{fig:dna_model}), and $\epsilon$ the free energy of duplexing. The force thus relies on the stability of DNA duplexes, which in turns depends on an externally controlled parameter: the temperature $T$. The higher the temperature, the less stable the duplex, and thus the less impact the functionalization will have on the swarm.

\subsection{C-GLASS discussion}

The interaction mechanisms introduced in the previous section have been added to the code of C-GLASS for the purpose of this work. %
We chose the proposed potential for two main reasons. %
First, we expect the model to have a spring-like mechanism, and second, it has to be proportional to the free energy of the model. %
Indeed, the more stable the structure, the less spurious opening of the duplex will happen, and thus the more we expect it to be rigid. %
However, a possible limitation of the proposed model is the lack of experimental data to back up the range values of the potential. %
A possible correction is to tune the potential with a multiplicative constant in Eq.~\ref{eq:potential} to match reasonable temperature ranges (e.g., $293.15$K$\sim313.15$K). %
However, as this work focuses on the semantic study of the simulation, the derivation of a proper constant is left as future work when \textit{in vitro} data become available. %
In this work, we use an arbitrary temperature range of $200$K to $400$K, selected for the observation of all the expected behaviors. %

Given the complexity of the simulator, many parameters intervene in the generation of the simulations. %
A full configuration file is available in the GitLab of the project\footnote{\href{https://gitlab.com/BACHARDT/SwarmRepLearn}{https://gitlab.com/BACHARDT/SwarmRepLearn}}. %
The main differences with the initial C-GLASS parameter file are the definition of the \textit{mt} (microtubule) object to model the swarms, and the switch in the potential evaluation to external temperature, which combines in a single file Lennard-Jones and our DNA potential.

In this work, we vary the temperature to check its impact on swarming behavior through its effect on the DNA duplexes. %
We  swipe over a temperature range from $200$K to $400$K, in $25$K steps. %
Each simulation lasts $500$ frames, and each frame is split into $9$ sub-frames in a grid pattern -- to better focus on specific behaviors. %
In total, we generated $40500$ frames for this work.

\subsection{Expected behaviors of the swarms}

As a way to evaluate the semantic model, we provide a list of behaviors that are expected to emerge in the system. Those will provide a ground truth with which to match the semantic atoms of the dictionary.
Be it simulation or \emph{in-vitro} experiments, microtubule swarms display several types of global behaviors. Moore \emph{et al.} provided 5 general types: (a) active isotropic, where microtubules move in a disorganized way, (b) flocking, where some microtubules remain alone, while others organize in small swarms, (c) giant flocking, where most of the microtubules are recruited by a single massive swarm, (d) spooling, where swarm are large enough and flexible enough to interact with themselves, forming stationary circular structures, and (e) swirling, where flexible swarms interact continuously crash into each other, creating large dynamical structures \cite{moore2020collective}. While the addition of DNA functionalization maintains the possibility for all of those structures \cite{KeyaKakugo2018}, we put particular emphasis on the group sizes, considering three categories (Figure \ref{fig:dna_model}, bottom): disorder (corresponding to the (a) category), partial swarming (corresponding to the (b) and (d) categories), and full swarming (categories (c) and (e)).

\begin{figure*}[htbp]
    \centering
    \hspace{2cm} Disordered \hfill \hspace{0.5cm} Partial swarming \hfill Strong swarming\hspace{2cm}
    
    \includegraphics[width=.33\textwidth]{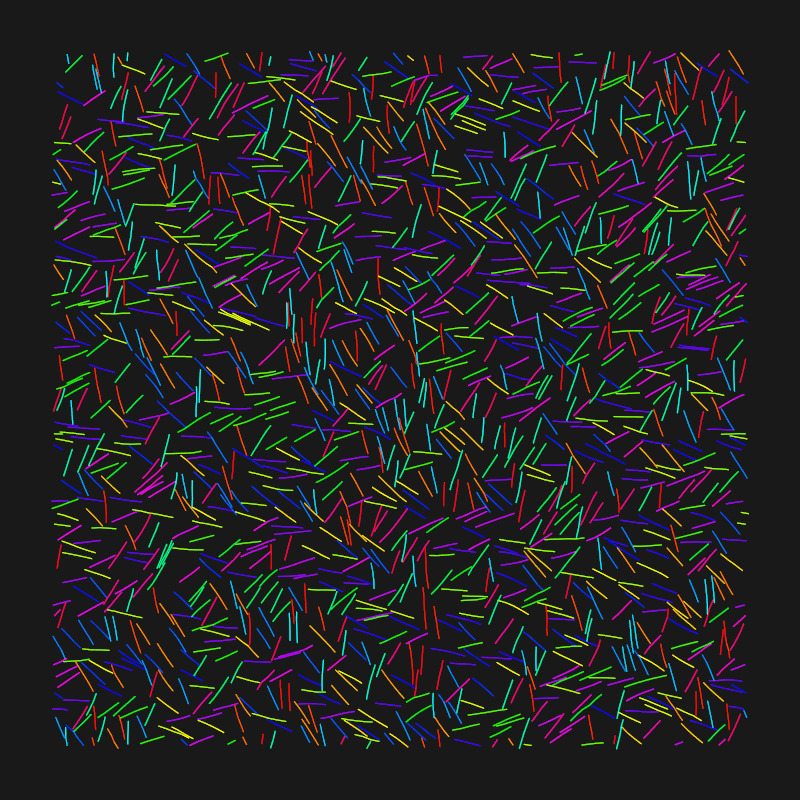}%
    \hfill%
    \includegraphics[width=.33\textwidth]{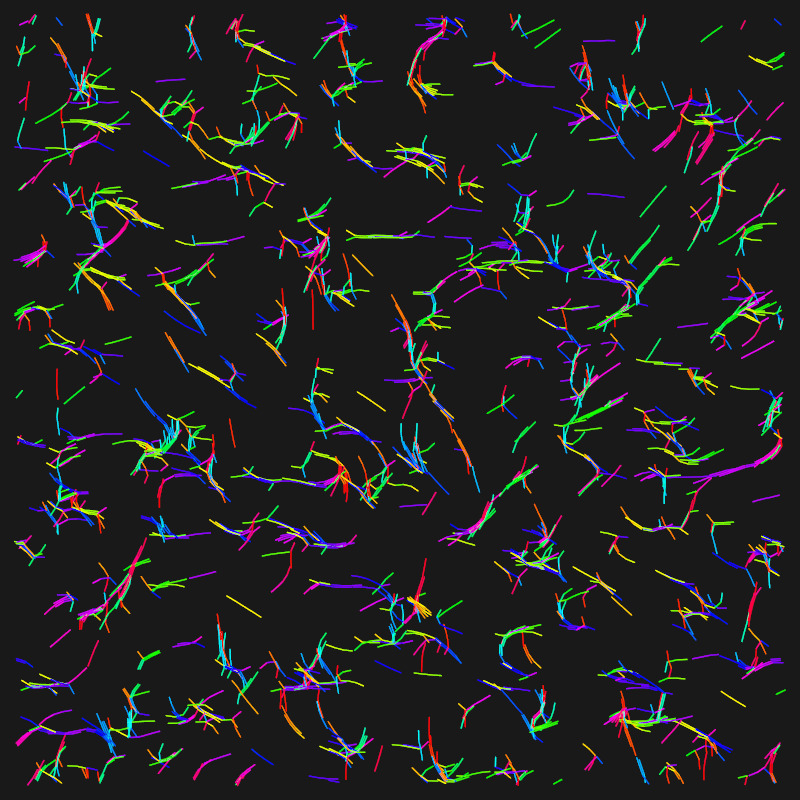}%
    \hfill%
    \includegraphics[width=.33\textwidth]{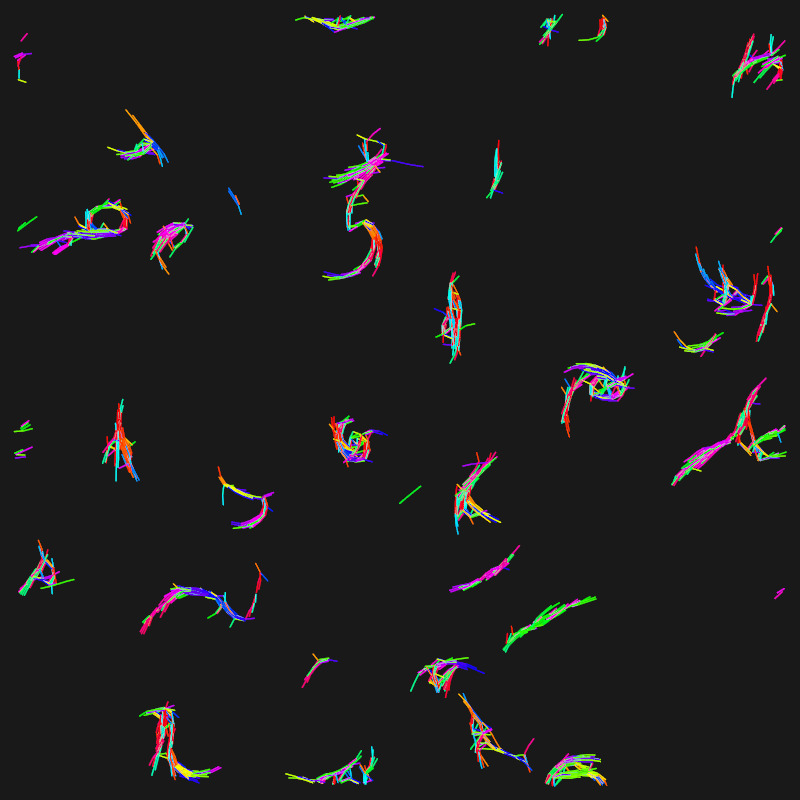}%

    \vspace{.08cm}

    \includegraphics[width=.33\textwidth]{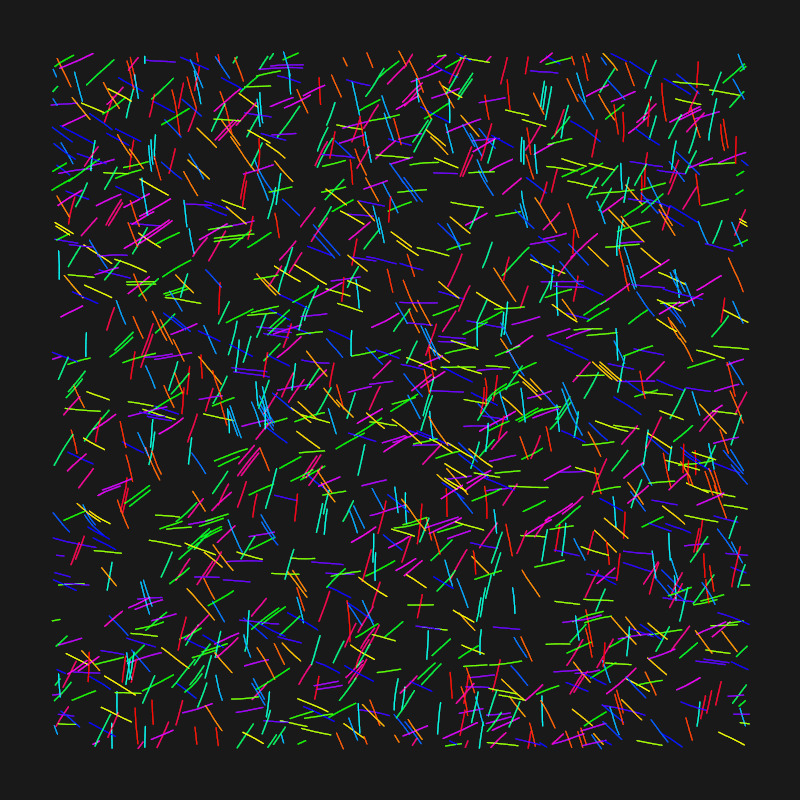}%
    \hfill%
    \includegraphics[width=.33\textwidth]{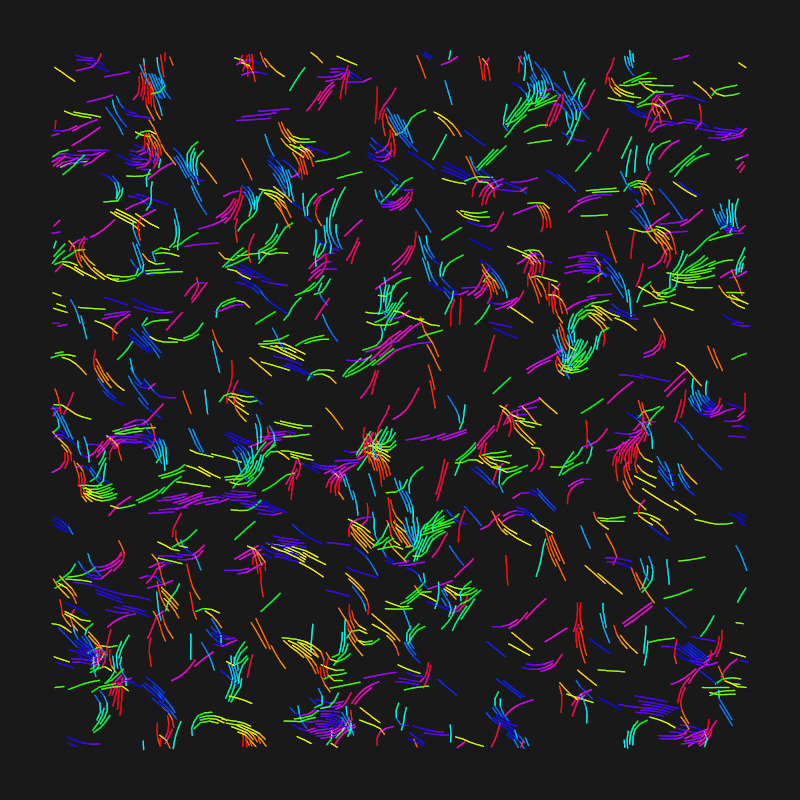}%
    \hfill%
    \includegraphics[width=.33\textwidth]{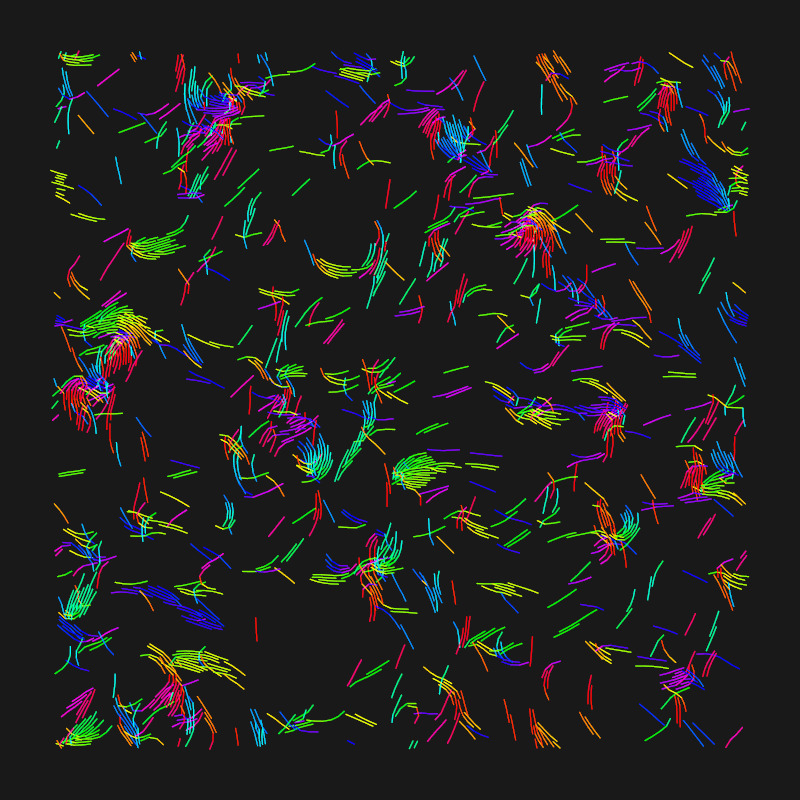}%
    \caption{Selected frames from the C-GLASS simulations with the given parameters. The color represents the angle of the microtubules relative to an arbitrary fixed reference. (Top to bottom, left to right) Random initialization common to all the simulations. Middle of the simulation for a temperature of $200$K. End of a simulation for a temperature of $200$K. End of the simulation for a temperature of $275$K. Middle of the simulation for a temperature of $400$K. End of a simulation for a temperature of $400$K.}
    \label{fig:ex_simul}
\end{figure*}

\subsection{\textit{In silico} simulations}

Given the interaction mechanisms introduced in C-GLASS with the aforementioned parameters, selected frames of the simulations are presented in Fig.\ref{fig:ex_simul} -- where the color of a microtubule represents its angle given an arbitrary fixed reference; this representation helps to notice structural patterns. %
As expected, we observe that the temperature indeed influences the behaviors of the simulated microtubules. %
Swarming appears at low temperatures and seems to be absent when the temperature is raised above $275$K -- arbitrary value, given the limitations of our approach. %
However, our capability for the characterization of behavior is limited to direct observations. %
Moreover, as the simulator simplifies the geometry of the \textit{in vitro} microtubules for computational necessities, the lack of visual resemblance of the simulations to the ground truth observation limits the possibility to conclude whether the simulations depict the correct behavior or not. %
To address this limitation, we propose a systematic approach towards linking the semantics of the simulated swarms to the expected behaviors observed in experiments in the rest of this paper. %

\section{Semantic embedding of the simulations}
\label{sec:embedding}

\subsection{Motivations}

By design, the simulations aim to resemble \textit{in vitro} experiments. To the naked eye, the behavior of the swarms seem to correctly mimics the behavior of experimental set-ups. %
However, due to the limitations of the C-GLASS simulator, the form and behavior of the simulated microtubules are not exactly the same as their \textit{in vitro} counterparts. %
Because of these differences, we introduce in this section a systematic method to confirm the presence of the aforementioned expected behaviors. %

The frames of a simulation can be considered as a collection of images depicting different interactions between the simulated microtubules. %
As a way to ensure that the simulations actually represent existing behaviors, we need a method providing strong classification elements (both for, that "is part of" a specific behavior, and against, that "cannot be part of" a specific behavior) to discriminate behaviors.
Given the semantic-behavior nature of this work, we propose to tackle this problematic through a semantic dictionary-based approach, based on \cite{bachard2024smic}.

\subsection{Semantic embedding}

\begin{figure*}[htbp]
    \centering
    \includegraphics[width=\textwidth]{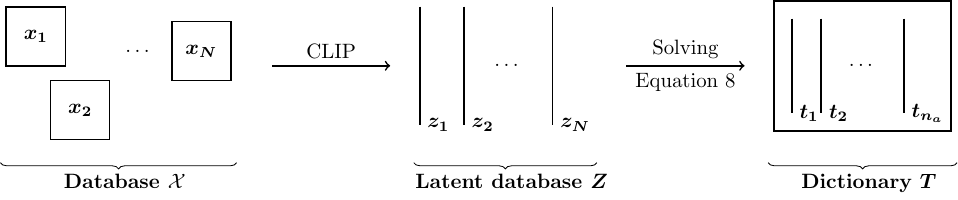}
    \caption{Learning a semantic dictionary from a given image collection $\mathcal{X}$. Pipeline from \cite{bachard2024smic}.}
    \label{fig:learn_dict}
\end{figure*}

In \cite{bachard2024smic}, the authors learn a dictionary based on a collection of images. %
The learning process, depicted in Fig.~\ref{fig:learn_dict}, is of a semantic nature. %
First, all the images are embedded into a semantic feature space, here using CLIP\cite{clip}, a foundation model. %
Then, we solve the minimization problem defined in Eq.~(\ref{eq:learn_dict}) to obtain a semantic dictionary, $\mathbf{T}^{*}=(\mathbf{t}_i)_{1\leqslant i \leqslant n_a}$. %
The components of this dictionary, the semantic atoms $t_i$, are a discriminative, non-redundant collections that span the semantic description of the image collection.

\begin{align}\label{eq:learn_dict}
    \mathbf{T}^{*} = \text{arg min} \frac{1}{2}\|\mathbf{Z}-\mathbf{T}\mathbf{C}\|^2_2 + \mu \|\mathbf{C}\|_1
\end{align}

where $\mathbf{Z}$ is the agglomerated (column-wise) matrix of the feature vectors returned by CLIP over the image collection, $\mathbf{T}$ is a dictionary, $\mathbf{C}$ is a matrix where each column represents coefficients associated to each atom of the dictionary for each images of the data collection and, and $\mu$ is the parsimony trade-off parameter, set to $1$ in this work. %

Using that approach, we learn a semantic dictionary over the previously introduced fixed temperature simulations.
Counting all the different experiments, a total of $40500$ images -- a good trade-off between expressivity and computation time\cite{bachard2024smic} -- is used to learn a semantic dictionary, made of $12$ atoms. %
This dictionary is used for all the semantic analysis introduced in the rest of the paper. %
    
\subsection{Visualizing the semantics of the simulations}

\begin{figure*}[htbp]
    \centering
    \includegraphics[width=\textwidth]{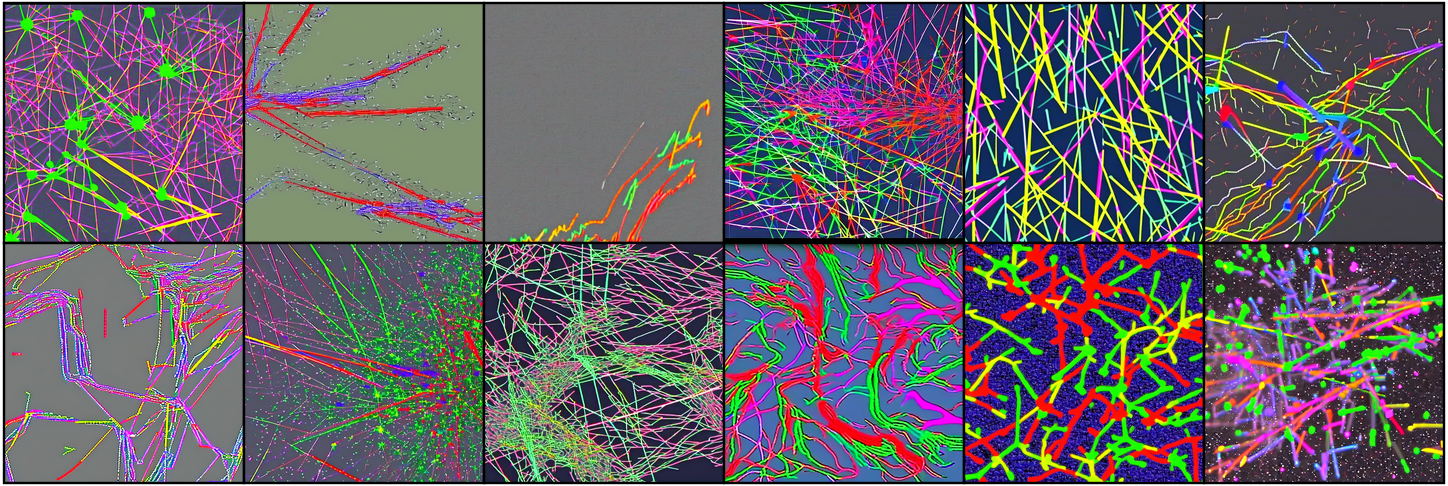}
    \caption{Sample images generated from the atoms of a semantic dictionary learned on the frame of the simulations. (Top to bottom, left to right) Ordered atoms by their relevancy (\textit{i.e.,} by their discriminant capacity) in the dictionary.}
    \label{fig:atom_simu}
\end{figure*}

The elements of the semantic dictionary are CLIP vectors. %
This property, inherited from \cite{bachard2024smic}, allows for the generation of natural images from these vectors using image generators, such as UnCLIP\cite{unclip}. %
Through the study of the images generated by this method, we claim that the learned dictionary encapsulates the semantics and the behaviors of the simulated swarms, accordingly to the expected behaviors described in Section~\ref{sec:model}. %
The generated images are presented in Fig.~\ref{fig:atom_simu}. %
Due to the stochastic nature of UnCLIP, we generate $5$ images of each atom and select the best images in terms of image quality, independently of the semantic content. %

A first general observation of the generated images shows that the atoms we learn on the data encapsulate the semantic behavior of the microtubules' swarms. %
In particular, we observe colored sticks either aggregating together or moving independently in an empty space, like in the simulation examples from Section~\ref{sec:model}. %
As such, we can conclude that a general representation of the simulations is encapsulated in the dictionary. %
However, the atoms do not perfectly encapsulate the geometric nature of the data: there is no clear delimitation of the microtubules nor smooth shapes, and colors do not exactly match the ones from the simulations. %
Nevertheless, those limitations should not impact the goal of this work, as we aim for interpretability rather than the ability to generate realistic swarm images. We are interested in the interactions between the objects (\textit{e.g.,} swarming and unswarming) rather than their individual behaviors, so that the faithful depiction of isolated microtubule is less impactful than not correctly depicting their macroscopic interactions. More generally, this geometric limitation of the CLIP-UnCLIP pipeline is already known in the literature\cite{clip, bachard2022clip} and does not necessarily interfere with the semantic nature of the results. %
Moreover, the models were not specifically trained to generate microtubules swarms -- as this specific use case was not present in the original training dataset\cite{laion} --, and we do not retrain nor fine-tune the models on our data. %

\begin{figure*}[htbp]
    \centering
    \includegraphics[width=0.33\linewidth]{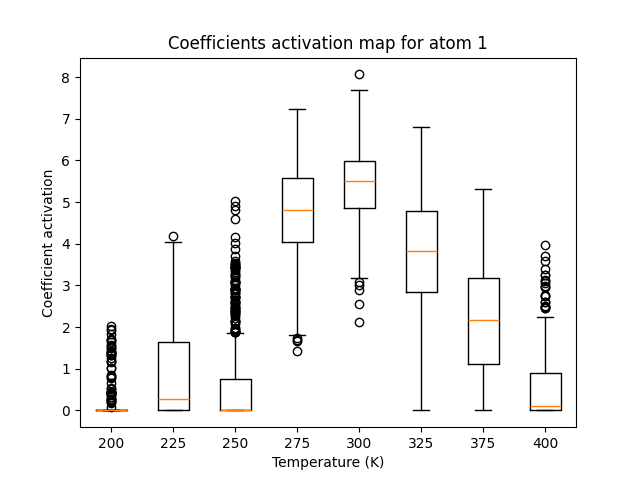}%
    \hfill%
    \includegraphics[width=0.33\linewidth]{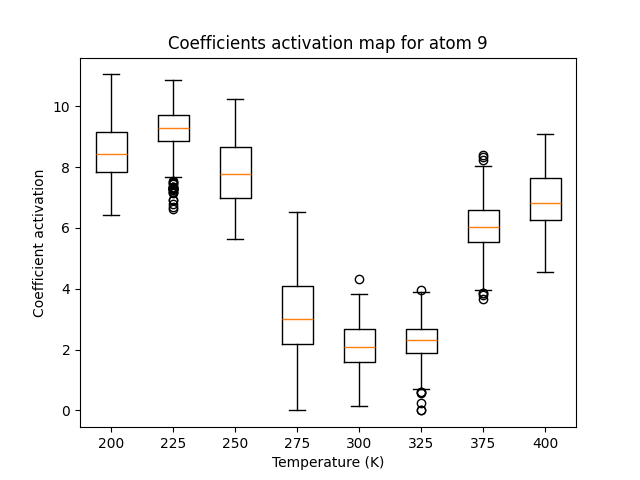}%
    \hfill%
    \includegraphics[width=0.33\linewidth]{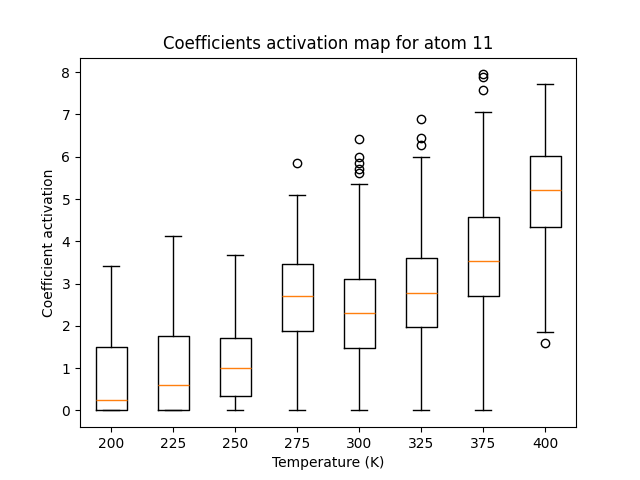}%
    \caption{Box plot activations for some atoms of the dictionary for the different temperatures of the simulations. The remaining atoms activation maps can be found in the appendix. The first $50$ frames of each simulation are not accounted in the computation as they do not represent typical behavior due to the non-physical random initialization.}
    \label{fig:atoms_map}
\end{figure*}

A more precise analysis of the images generated from the atoms of the semantic dictionary allows us to better see that this approach encapsulates different behaviors according to the experimental set-ups. %
From the natural images generated in Fig.~\ref{fig:atom_simu}, we observe different swarm densities. %
These differences can be assimilated to different phases of swarming, -- \textit{e.g.,} low density indicates strong swarming while high density indicates little to no swarming. %
That intuition is strengthened by a more systematic approach depicted in Fig~\ref{fig:atoms_map}.
This figure shows, for different atoms, typical activation values for different values of the temperature. %
For example, we can observe that atom $1$ mostly activates during non-swarming phases, at temperatures around $325K$, while, on the other hand atom $9$ mostly activates at temperature where strong swarming is expected (either at low temperatures, or when the simulations do not follow physical laws at excessive temperatures). %
The activation maps for the remaining atoms are presented in the appendix where other behaviors can be observed. %

Through these observations, we show that the dictionary based approach seems to correctly extract relevant atoms from the simulation to classify the behaviors of the swarms. %
Indeed, without any external information, this approach is capable of finding the expected behaviors, based on the external temperature. %
To further strengthen this claim, the next section looks up the decomposition of frames in this dictionary. %
Indeed, as multiple atoms activate for a frame, this decomposition comes in handy to study the relevancy of the proposed atoms. %

\section{Semantic interpretation of the simulations}
\label{sec:interpretation}

\subsection{Semantic decomposition of frames of the simulation}

We demonstrated in the previous section that we can extract relevant semantic behaviors from the microtubules' data and combine them into a semantic dictionary.
In this section, we are interested in the decomposition -- \textit{i.e.,} which atoms are activated, and the values associated to these activations -- of the frames over the learned semantic dictionary.
The goal of this work is to show that only the expected atoms are activated in each of the different simulation set-ups.

To obtain the semantic decomposition of a frame into the dictionary, we solve the minimization problem depicted in Eq.~(\ref{eq:learn_coef}). %
This equation is adapted from Eq.~(\ref{eq:learn_dict}) for a unique latent vector $\mathbf{z}=\text{CLIP}(\mathbf{x})$, the frame we aim to decompose, and we solve for $\mathbf{c}$ the coefficients values, instead of $\mathbf{T}$ the dictionary. %

\begin{align}\label{eq:learn_coef}
    \mathbf{c}^{*} = \text{arg min} \frac{1}{2}\|\mathbf{z}-\mathbf{T}\mathbf{c}\|^2_2 + \mu \|\mathbf{c}\|_1
\end{align}

\begin{figure*}[htbp]
    \centering
    \includegraphics[width=\textwidth]{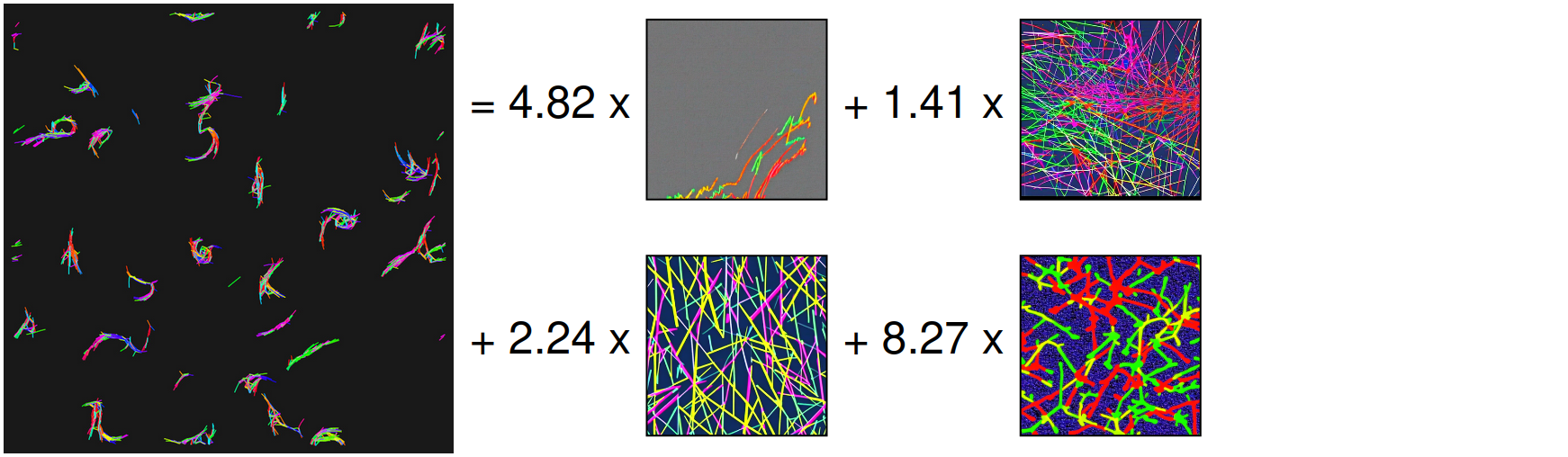}
    
    \hspace{.1cm}
    
    \includegraphics[width=\textwidth]{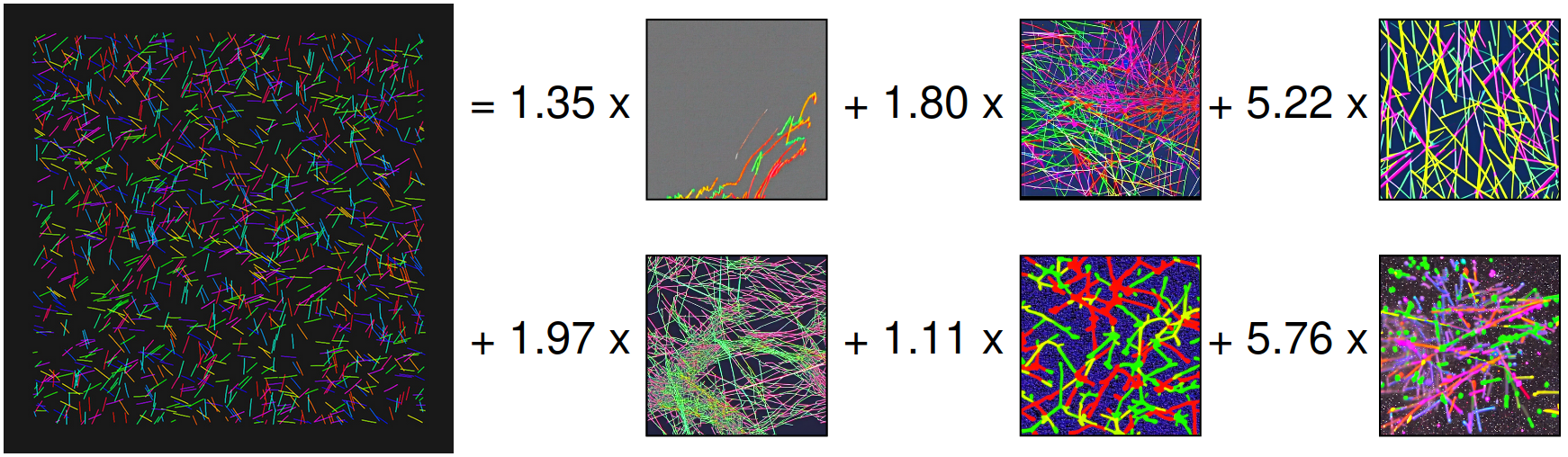}
        
    \hspace{.1cm}
    
    \includegraphics[width=\textwidth]{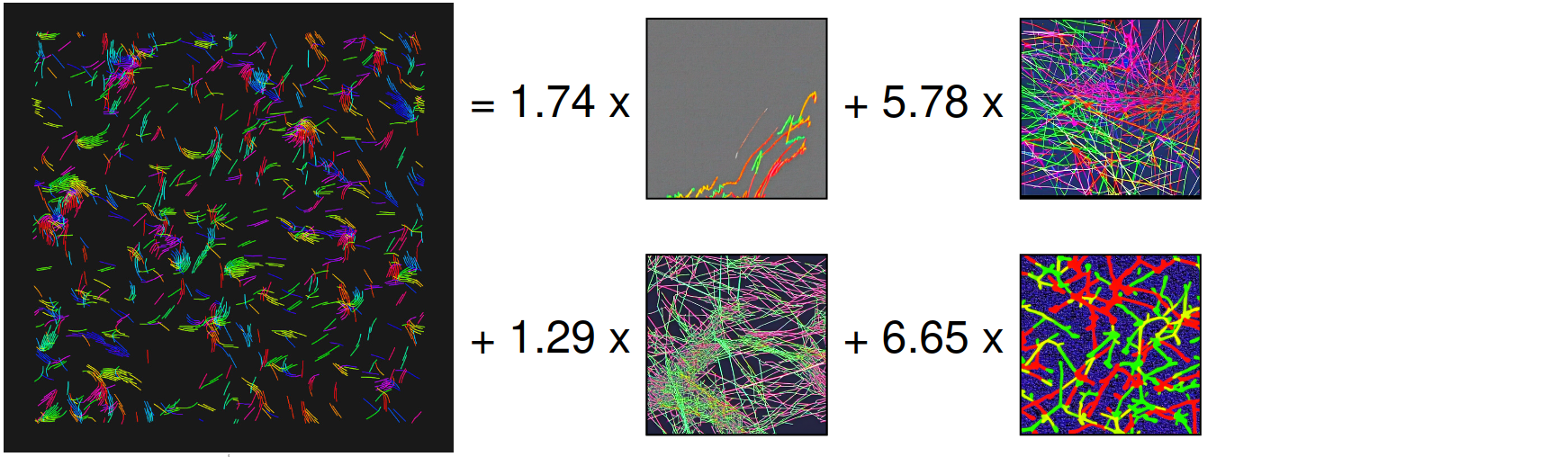}
        
    \caption{Semantic decomposition of frames in the dictionary. (Top to bottom) Initial random frame common to all the simulations. Final frame of the simulations for a temperature of respectively $200K$, $300K$ and $400K$.}
    \label{fig:frame_decomposition}
\end{figure*}

A qualitative approach for the decompositions of atoms in the dictionary is presented in Fig.~\ref{fig:frame_decomposition}. %
The proposed examples are the decomposition of relevant frames of the simulations: the first frame, common to all the simulations, and the last frame of simulations for a temperature of $200K$, $325K$ and $400K$. %
From the selected examples, we first observe that some of the atoms are common to all the decomposition. %
Indeed, as the semantic space spanned by CLIP is huge, and because the simulations depict the same system, we ought to find a consistent decomposition basis. %
However, the decompositions are not identical in terms of activated atoms, and the values associated to the activated atoms also vary. %
We postulate that the semantic differences between the behaviors of the simulated swarms lie in these subtle variations.

\begin{figure*}[htbp]
    \centering
    \includegraphics[width=0.33\linewidth]{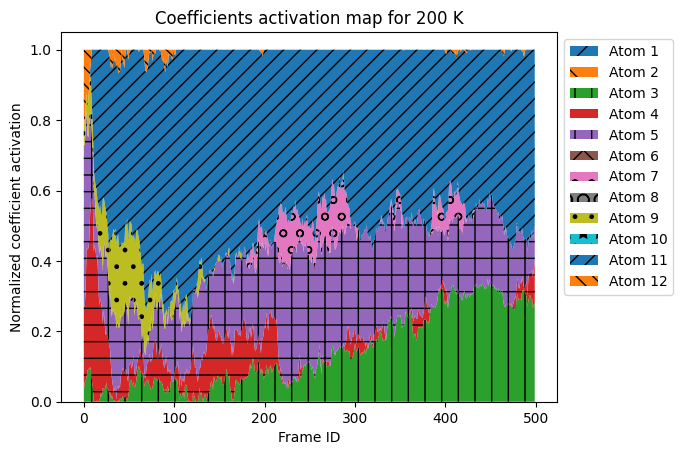}%
    \hfill%
    \includegraphics[width=0.33\linewidth]{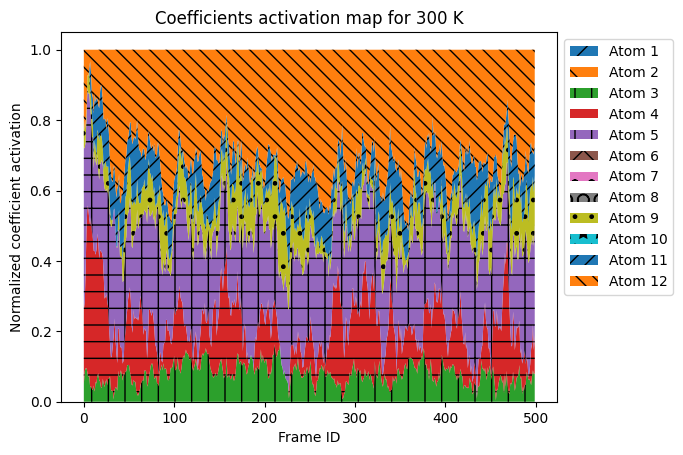}%
    \hfill%
    \includegraphics[width=0.33\linewidth]{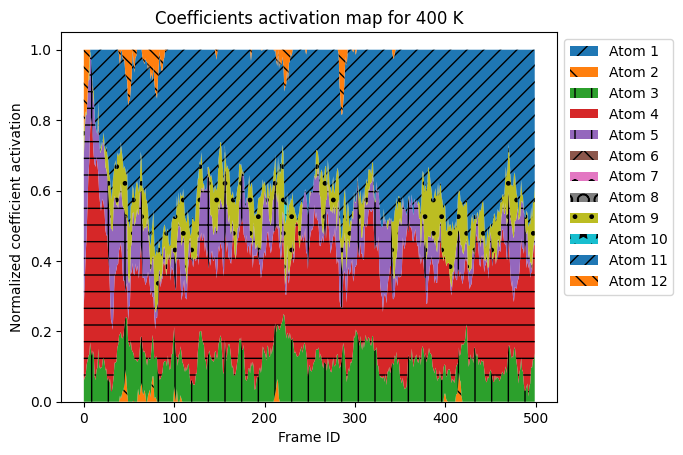}
    \caption{Temporal evolution of the atoms' activations for different experiments. (Left to right) Temperature respectively set to $200K$, $300K$ and $400K$.}
    \label{fig:activation_maps_temp}
\end{figure*}

Indeed, if we observe the evolution of the activation values of the atoms as the simulations progress, we observe very different dynamics. %
These activation maps for a temperature of $200K$, $325K$ and $400K$ are presented in Fig~\ref{fig:activation_maps_temp}. %
From these activation maps, we can see that the presence, or absence, of swarming induces very different dynamics, meaning that the semantic approach can distinguish the expected behaviors. %
More interestingly, this approach classifies the expected behaviors without prior knowledge of the system, and without retraining. %
This result motivates us to use semantics to infer information from the system, such as for example the temperature of the system. %

Another observation from these decompositions, which is strengthened by the results depicted in the supplementary material, is the fact that only a few atoms encode most of the relevant behaviors. %
Indeed, we can hypothetize that the semantic dictionary mainly encapsulates high level semantics behavior, in the same idea as the behavior depicted in Sec.~\ref{sec:model}. %
However, it is yet to say whether there is a one to one correspondence between the atoms of the dictionary and the expected behaviors. %
The study of this idea is outside the scope of this work and left as future work. %

\subsection{Inferences through semantics}

As depicted in the figures in the previous section, the semantic learned over the simulations depict different profile regarding the values of the external temperature. %
We demonstrate here with a proof-of-concept experiment that, given only the activation values of the atoms, we are able to estimate the temperature of the simulation. %

\begin{figure}
    \centering
    \includegraphics[width=\linewidth]{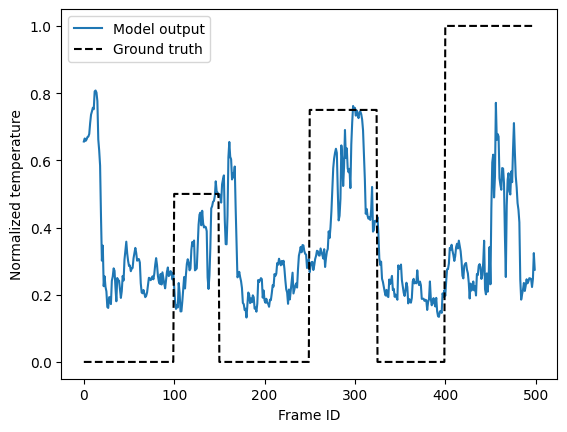}
    \caption{External temperature estimation using the semantic activation for each frame.}
    \label{fig:res_learning}
\end{figure}

We consider a relatively small multi-layer perceptron -- 2 layers, $\sim500$ parameters in total  -- that take as inputs the activation values for a given frame and outputs the current temperature of the simulation. %
We train this model by learning the expected temperature on the same dataset as we used to learn the semantic dictionary. %
Then, we generate a new simulation, where the temperature changes over time, and use the activation values for each frame to estimate the current temperature. %
An example of the results of the learning is given in Fig.~\ref{fig:res_learning}. %

From the results, we observe that for all the parameters sets, the model is able to correctly predicts the expected temperature of the simulations. %
We also observe that the model has a slight delay before adapting to a sudden temperature change, which is consistent with expected molecular behaviors.
Indeed, during \textit{in vitro} experiments (\emph{e.g.}, \cite{KeyaKakugo2018, kawamata2024autonomous}), the same phenomenon is present: the swarms do not instantly swarm or un-swarm when the control parameter changes, a short delay is present. %
This result shows that meta-information of the swarms -- such as hyperparameters -- are encoded in the semantic dictionary, showing a good understanding of the behavior of the system. %

These observations are coherent with the evaluation of the learning task: over ten repeats, we obtained an MSE of $0.22\pm 0.04$. %
This low values with a low standard deviation  indeed shows that the model is able to learn the (normalized) temperature of the simulations given only the activations values of the semantic atoms of the simulation. 

Of particular interest, we note that some of the errors come from a lag in the estimation when quickly switching temperature. That lag is consistent with the observed behavior (both \emph{in-silico} and \emph{in-vitro}) where swarms will take some time to disassemble, microtubules at the periphery leaving first, progressively leaving space for more central elements to separate. Similarly, when the conditions become favorable to swarming (low temperature), a lag is induced by the time necessary for microtubules to collide and assemble. 

Our results also highlight room for improvements. %
Indeed, the model does not seem to be able to output a relative temperature of $0$ -- when the swarming is the strongest -- or sometime overshoots the relative temperature when the unswarming effect is not the strongest. %
These results also highlight the limits of the current proof-of-concept approach; more adapted and complex models may be able to achieve a better link between the semantic atoms and the external temperature of the simulations. %

\section{Conclusion}

Overcoming the shortness of experimental data by simulating the behavior of the studied material is common practice in biology. %
However, evaluating the correctness of behavior in such simplified systems is challenging. %
In this work, we showed that the proposed DNA-functionalized microtubule swarm simulator, based of C-GLASS, exhibits the expected macro behaviors of their \emph{in vitro} counterpart -- namely swarming, separating, and the in-between processes. %
To strengthen this claim, we extracted the most discriminating semantic descriptors of the experimental simulations by learning a semantic dictionary of swarm behaviors. %
We showed that the extracted atoms matched the expected behaviors. %
Moreover, we showed that the decomposition of each frames of the simulations correctly describes the expected behavior of the swarm regarding external parameters values. %
This result provides relevant leads towards the explainability of simulated experiment, providing a first step towards automated design, evaluation and optimization. %

This work opens the way to several future perspectives. %
First, in this work, we limit the study to \textit{in silico} data. %
The natural next step is to apply the proposed approach to \textit{in vitro} experimental data. %
Furthermore, we could then study the links between the experimental and simulated semantic dictionaries. %
This could be a serious lead to overcome the lack of experimental data by providing an approach for the generation of realistic synthetic data. %

However, the current lack of realism of the simulations is an obstacle to linking the two approaches. %
Technical enhancements of the simulator are thus an interesting venue for future work. %
Indeed, a more precise and complex model, or  more realistic frames --\textit{e.g.,} graphic redesign or preprocessing layers --, would benefit the realism of the \textit{in silico} experiments. %
With better simulations, finding a connection between experiments and simulations would be less challenging. %
We could use this link as a metric to quantify to which extent the simulations resemble experimental data, and use this information to continuously improve the simulator. 

Further, a more data driven application of this work would be to use the activations values of the atoms as inputs for a learning task. %
Indeed, as the atoms encapsulate a high level description of the behavior of the swarms, they are appropriate inputs for machine learning tasks, such as processing data encoded as external parameters values. %

Finally, the fact that only a few atoms -- $12$ in this work -- are apparently needed to encapsulate the behavior of the simulations also suggests that this approach could be used in frameworks where the learning capabilities are limited, including embedded learning or reservoir computing. %

\section*{Acknowledgements}

This work was supported by JSPS KAKENHI Grant Numbers JP25H00608 and 25K00151. IK was supported by the Sumitomo Foundation, the Foundation of Kinoshita Memorial Enterprise, and the Iketani Science and Technology Foundation.

\bibliographystyle{plain}
\bibliography{references}

\newpage

\onecolumn

\appendix 

\section*{III. Semantic Embedding of the Simulations}

\begin{figure*}[h!]
    \centering
    \includegraphics[width=0.33\textwidth]{figures/atom_activations/atom_1.png}%
    \hfill%
    \includegraphics[width=0.33\textwidth]{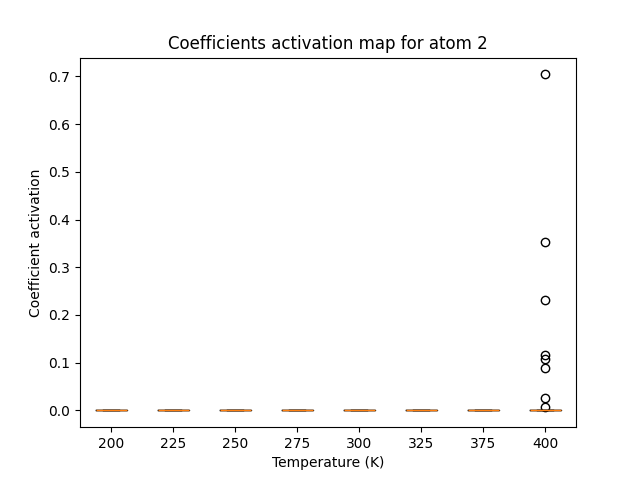}%
    \hfill%
    \includegraphics[width=0.33\textwidth]{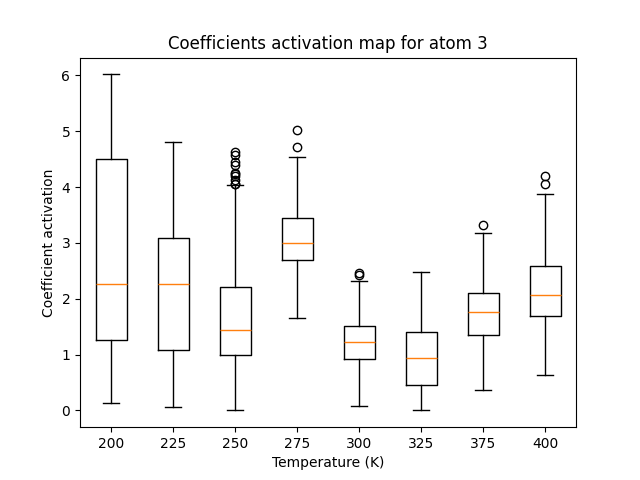}%
    
    \includegraphics[width=0.33\textwidth]{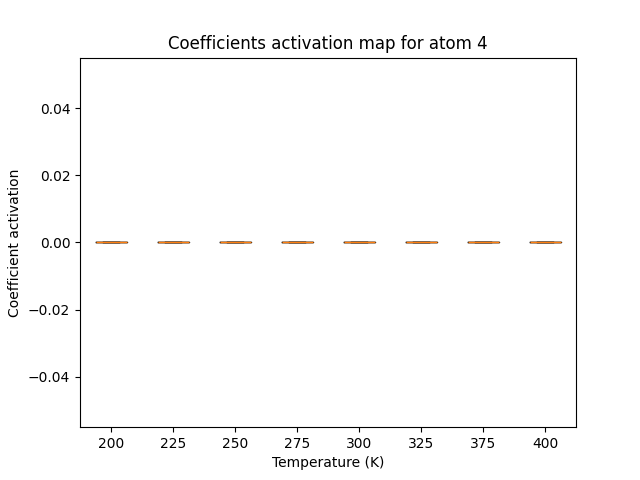}%
    \hfill%
    \includegraphics[width=0.33\textwidth]{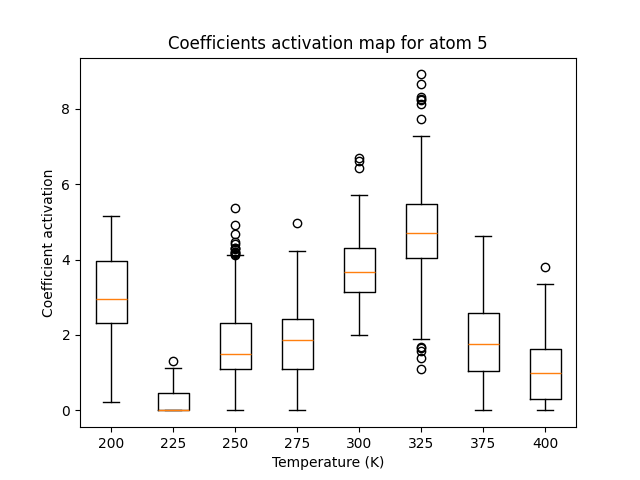}%
    \hfill%
    \includegraphics[width=0.33\textwidth]{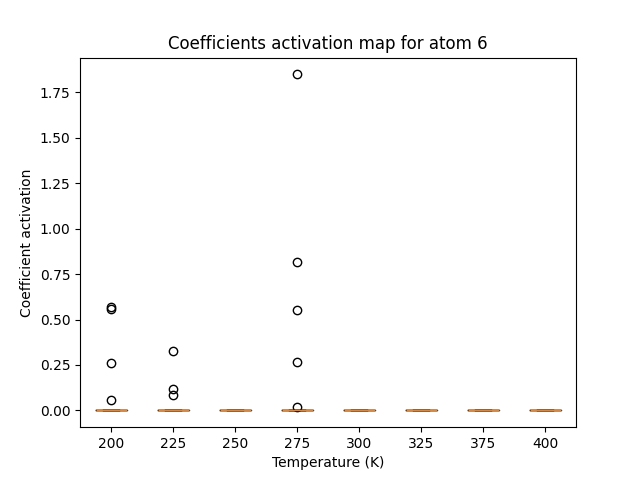}%
    
    \includegraphics[width=0.33\textwidth]{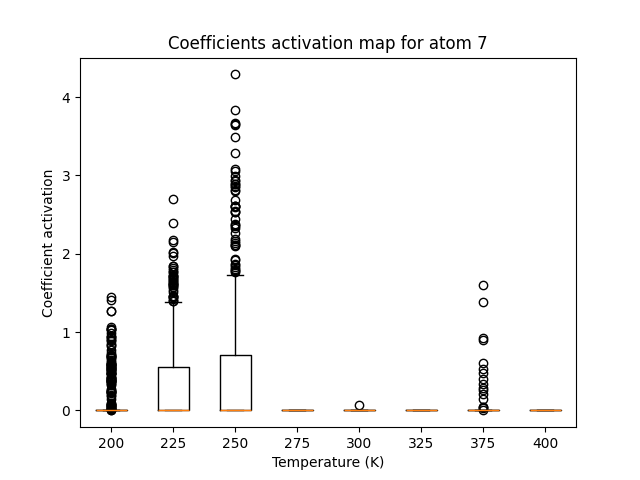}%
    \hfill%
    \includegraphics[width=0.33\textwidth]{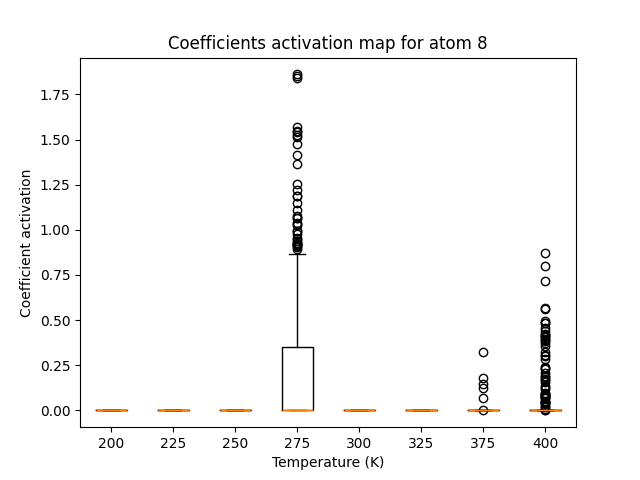}%
    \hfill%
    \includegraphics[width=0.33\textwidth]{figures/atom_activations/atom_9.png}%
    
    \includegraphics[width=0.33\textwidth]{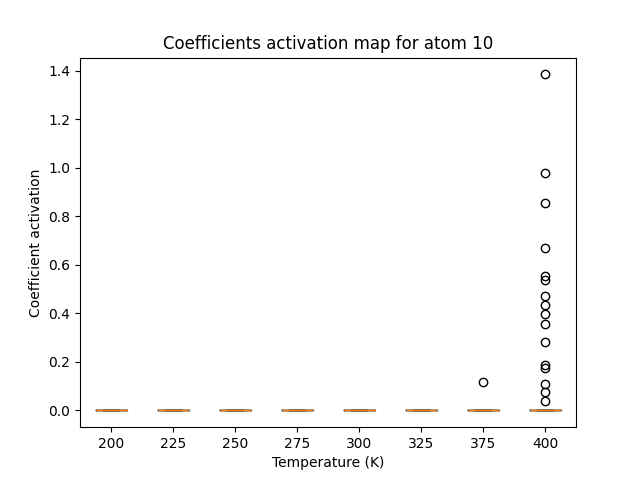}%
    \hfill%
    \includegraphics[width=0.33\textwidth]{figures/atom_activations/atom_11.png}%
    \hfill%
    \includegraphics[width=0.33\textwidth]{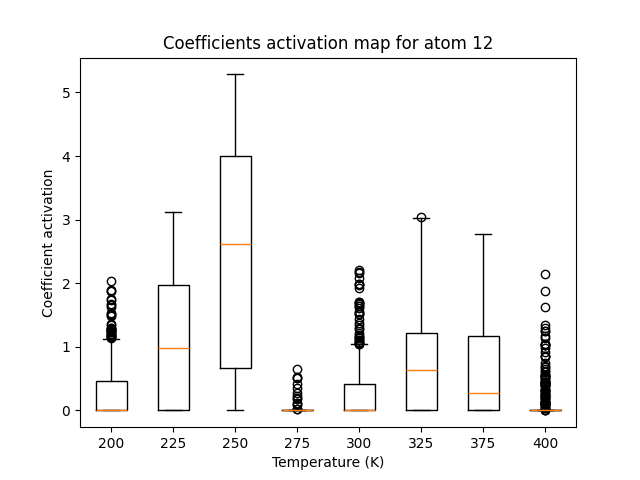}%
    
    \caption{Box plot activations for all the $12$ atoms of the dictionary for the different temperatures of the simulations. The $50$ first frames of the simulations are not accounted as they only describe the initialization of the simulations rather than the studied permanent regime.}
    \label{supp:atomes_activation}
\end{figure*}

\newpage

\section*{IV. Semantic Interpretation of the Simulations}

\begin{figure*}[h!]
    \centering
    \includegraphics[width=0.33\textwidth]{figures/temp_activations/temp_200.png}%
    \hfill%
    \includegraphics[width=0.33\textwidth]{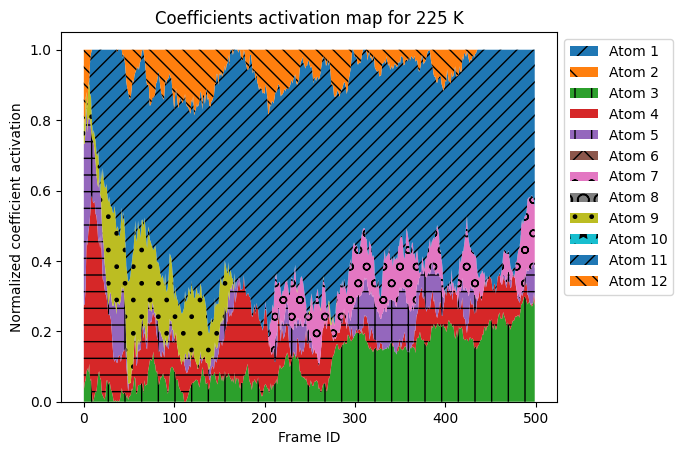}%
    \hfill%
    \includegraphics[width=0.33\textwidth]{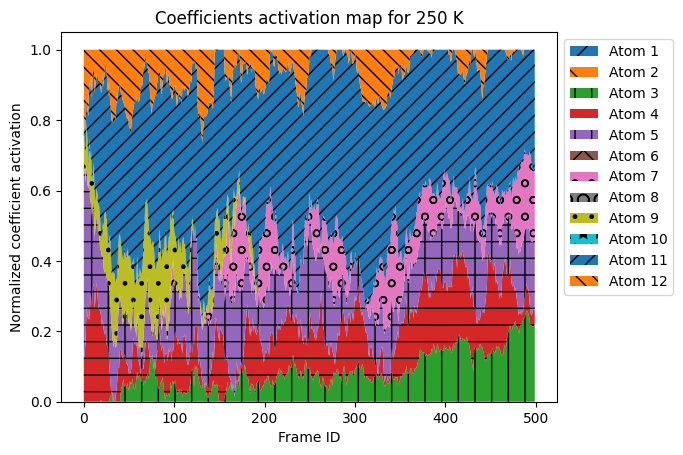}%
    
    \includegraphics[width=0.33\textwidth]{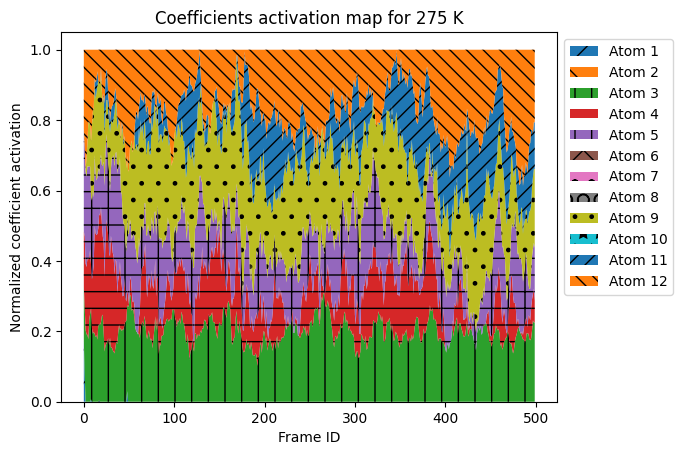}%
    \hfill%
    \includegraphics[width=0.33\textwidth]{figures/temp_activations/temp_300.png}%
    \hfill%
    \includegraphics[width=0.33\textwidth]{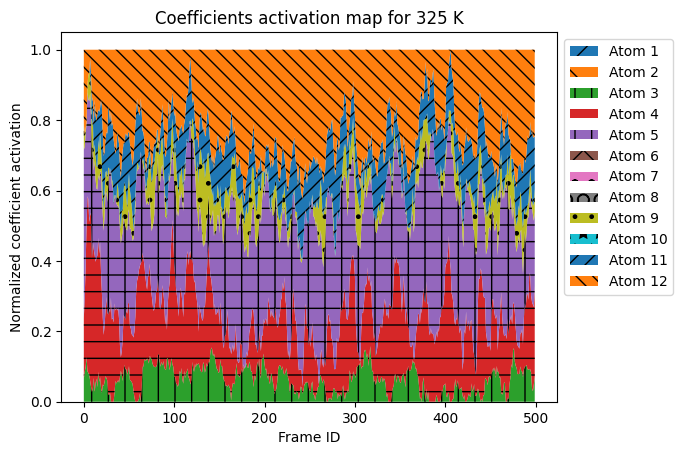}%
    
    \includegraphics[width=0.33\textwidth]{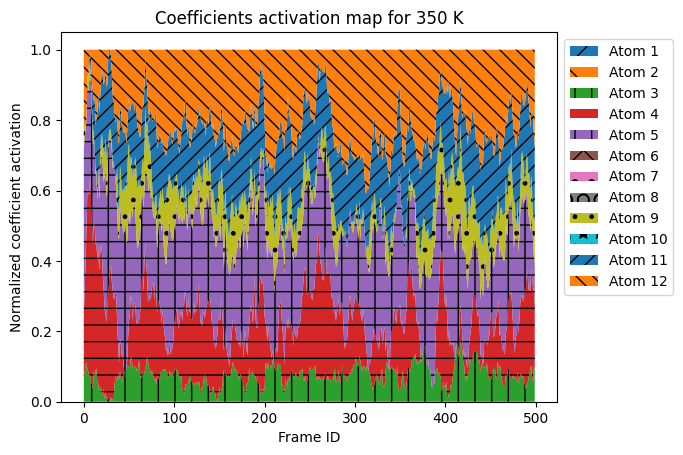}%
    \hfill%
    \includegraphics[width=0.33\textwidth]{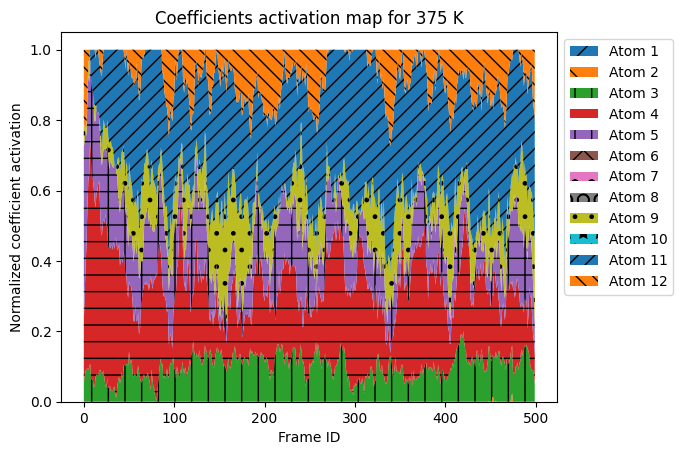}%
    \hfill%
    \includegraphics[width=0.33\textwidth]{figures/temp_activations/temp_400.png}%
    
    \caption{Temporal evolution of the activations of the atoms for all the different temperatures set-ups. From top to bottom, left to right, the temperature is $200 + i\times25K,\ i\in[0,...8]$.}
    \label{fig:placeholder}
\end{figure*}

\end{document}